# Self-supervised training for high-resolution close-range multispectral remote sensing imagery

**Leon-Friedrich Thomas[1,2], Mikael Änäkkälä[1], and Antti Lajunen[1]**

[1] University of Helsinki, Department of Agricultural Sciences, P.O. Box 28, Fi-00014 Helsinki, Finland
[2] University of Helsinki, Department of Forest Sciences, P.O. Box 27, Fi-00014 Helsinki, Finland

Corresponding author: Leon-Friedrich Thomas (e-mail: leon-friedrich.thomas@helsinki.fi).

"This project received funding from the Maatalouskoneiden Tutkimussäätiö (Agricultural Machinery Research Foundation). Open access funded by Helsinki University Library."

**ABSTRACT** Although self-supervised learning (SSL) offers a promising way to reduce annotation effort in close-range remote sensing, its effectiveness for high-resolution multispectral unmanned aerial vehicle (UAV) imagery remains underexplored due to limited data. This study evaluated SSL pretraining for precision agriculture using cm-scale multispectral drone imagery collected across multiple sensors, years, and regions. Transformer-based encoders were pretrained with Momentum Contrast v3 (MoCo-v3) and Masked Autoencoders on a harmonized dataset combining msuav500K with newly collected multi-year UAV imagery from agricultural fields in Finland. Pretraining used four spectral bands (Green, Red, Red-Edge, Near-Infrared) for cross-sensor compatibility. The models were evaluated on crop-weed semantic segmentation using the WeedMap dataset with 5–100 \% training data. The following two subsets served as downstream tasks: Task A (Germany, RedEdge-M), where all pretrained models were compared under partial and full fine-tuning, and Task B (Switzerland, Sequoia), where the best encoder from Task A was assessed. Our Swin Transformer pretrained with MoCo-v3 achieved the strongest performance on both tasks, surpassing the Swin Transformer model of Doornbos et al. pretrained on a pre-release of msuav500K. Our pretrained Swin Transformer further demonstrated cross-sensor and cross-region generalization. We additionally provide a public multi-year multispectral UAV dataset from Finland to support future research.

## I. INTRODUCTION

In recent years, deep learning has received significant attention in the field of remote sensing and has become increasingly integrated into precision agriculture research. This trend leverages the advancements in computer-vision models, particularly in tasks such as weed detection. While fully supervised deep learning and transfer learning are the predominant techniques for training and deploying deep-learning models, recent progress in computer vision have led to development of numerous foundation models tailored to remote-sensing applications [1]. These models are typically trained in a self-supervised manner on sensor-specific data or across a select range of sensor datasets [1]. In this study, we pretrained deep-learning models using high-resolution multispectral unmanned aerial vehicle (UAV) imagery acquired across multiple years, sites, and sensor types, suitable for precision agriculture. The models were pretrained in a self-supervised manner using a Momentum Contrast (MoCo-v3) or Masked-Autoencoder (MAE) framework on a large-scale multispectral dataset msuav500K, which was extended and harmonized with a self-collected multi-year multispectral UAV dataset of agricultural fields in Finland [2-4]. This combined dataset spans multiple countries, environments, and a variety of multispectral sensors used in precision agriculture and forestry. As no Nordic countries were previously represented in the msuav500K dataset due to lack of publicly available data, the combined and extended dataset which is utilized in this research is referred to as

msuav500k+N [2]. To evaluate the pretrained models, we utilized a downstream task of crop weed detection using the publicly available WeedMap dataset with ground sampling distance (GSD) 1 cm, which was collected with common multispectral sensors frequently used in UAV research [5]. WeedMap consists of two datasets that we used as two different downstream tasks. One was collected in Germany using a RedEdge-M sensor, which we used as downstream task A and evaluated all pretrained models on it [5]. The other was collected in Switzerland using a Sequoia sensor, which we used as downstream task B, on which we only evaluated the best-performing encoder of downstream task A [5].

As part of this work, we publicly released a subset of the self-collected and harmonized multispectral UAV image tiles at 0.7 cm GSD with 512 × 512 pixels and 4–5 spectral bands [6]. This release addresses the current lack of openly available high-resolution multispectral UAV data from Nordic regions, particularly Finland [2].

Self-supervised pretraining was conducted using only the Green, Red, Red-Edge, and Near-Infrared (NIR) spectral bands to maximize sensor compatibility. This selection aligns with the four- and five-band sensors in the dataset and reflects common multispectral configurations in precision agriculture, where the Blue band is frequently unavailable. By restricting the input to these bands, the pretrained models, which are intended for public release, remain broadly applicable across a wide range of sensors and platforms. In this study we addressed the following hypotheses:

**(i)** Self-supervised pretraining on the combined msuav500k+N dataset improves semantic segmentation performance on downstream Task A, particularly in low-data regimes, compared with (a) sensor- and geography-specific models trained end-to-end from scratch and (b) models initialized with ImageNet-pretrained weights. Evaluation is conducted on the WeedMap dataset collected in Germany using the RedEdge-M sensor.

**(ii)** The best-performing self-supervised pretrained model trained on msuav500k+N demonstrates cross-geographical and cross-sensor generalization, achieving competitive performance on downstream Task B using the WeedMap dataset collected in Switzerland with the Sequoia sensor.

**(iii)** The best-performing self-supervised pretrained model trained on msuav500k+N outperforms the Swin Transformer trained on msuav500k by Doornbos et al. (2025), which was based on a pre-release multispectral-only version of the dataset.

The main contributions of this work are:

1. **Self-supervised pretraining on multispectral imagery for precision agriculture.** We trained self-supervised models using high-resolution multispectral UAV imagery collected across multiple sensors, years, and geographic regions.

2. **Extended-dataset pretraining and performance gains.** We trained the first models on an extended version of msuav500K, expanded with additional high-resolution multispectral imagery from Nordic agricultural regions. The extended dataset, referred to as msuav500k+N, includes green, red, red-edge, and NIR bands at cm-level spatial resolution. The resulting pretrained Swin Transformer outperformed a counterpart trained solely on the pre-release msuav500K dataset by Doornbos et al 2025 [7].

3. **A public release of harmonized multispectral UAV dataset at 0.7 GSD**, providing previously unavailable data from Nordic agricultural regions, specifically from Finland.

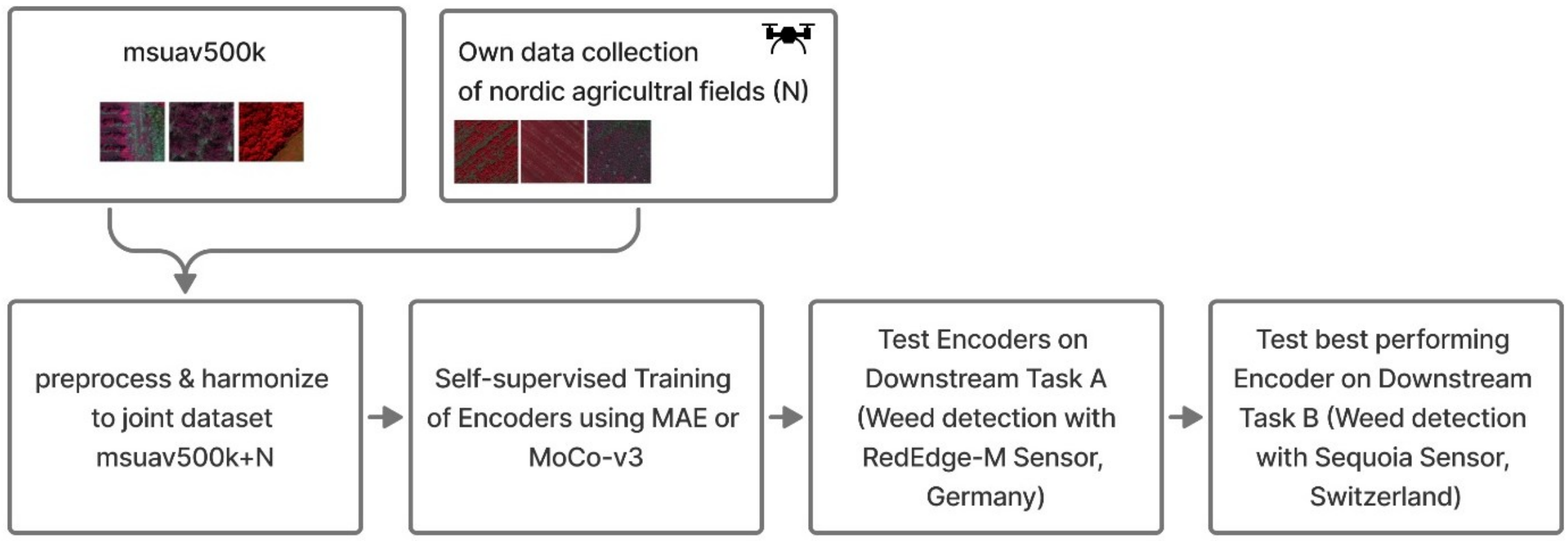

**Figure 1:** Summary Workflow of research

## II. Related Work

### A. Self-supervised Learning

Deep-learning models are widely used in remote-sensing research, including applications in precision agriculture, where weed detection in crop fields is a common task. In recent years, self-supervised learning (SSL) has become increasingly popular for representation learning in remote sensing [1]. Momentum Contrast (MoCo), particularly MoCo-v2, and MAE are among the most widely used SSL methods and have been applied in several foundation model pretraining studies [3, 8-12]. Other SSL approaches, such as DINO (v1–v3) and SimCLR, are also frequently used for foundation model pretraining [9-11].

These SSL methods are commonly applied to pretrain encoder backbones on large-scale remote-sensing datasets from sensors such as Sentinel-1/2, Landsat, and EnMAP [9-11]. Due to the large scale of satellite data archives, these datasets often comprise millions of images; for example, Landsat-based pretraining is typically conducted on several million images per sensor.

Common backbone architectures used in foundation model pretraining include ResNet-18, ResNet-50, and basic Vision Transformers (ViT), providing both convolutional and transformer-based baselines [9-11]. While different SSL strategies have been explored, MoCo and MAE have shown competitive performance in previous SSL experiments [3, 9-12]. Most foundation models are pretrained on sensor-specific datasets, focusing on a single sensor such as Landsat or Sentinel-2[1]. More recently, multimodal foundation models, such as TerraMind, Prithvi, and Copernicus-FM have been introduced, combining data from multiple sensors and modalities, pretrained using MAE-based and multimodal masked-modeling approaches [3, 13-15].

In related work, Doornbos et al. applied FastSiam for a short training schedule to train Resnet18, Resnet50, and Swin Transformer on the msuav500k dataset, which consists of multispectral and RGB drone imagery [2]. Unlike satellite-based datasets, msuav500k does not provide global coverage, as achieving such coverage with drones is considerably more challenging. However, the dataset spans multiple sensors with similar spectral bands and diverse geographic and environmental settings, making it suitable for learning transferable representations from high-resolution aerial imagery [2].

### B. Precision agriculture

Precision agriculture is a rapidly advancing field that increasingly relies on deep learning to enable data-driven farm management. Its primary objectives include monitoring crop and livestock health, optimizing production systems, and supporting timely interventions based on plant physiological conditions, thereby fostering the development of smart, technology-driven farming practices [16]. The growing use of UAVs and ground-based vehicles enables efficient, high-resolution data collection for decision-support systems, supporting applications such as crop disease detection, nutrient management, and weed monitoring, as well as targeted actions including selective fertilization and precision spraying [16, 17].

Weed detection is a key application of precision agriculture, as weeds directly reduce crop yield while increasing production costs and environmental impacts [18]. Automated, accurate weed detection enables site-specific management and selective spraying, reducing herbicide use and mitigating negative effects on ecosystems and human health, which are essential to ensure sustainable food production and to increase crop yield [18].

Ground-based robotic platforms and agricultural vehicles have seen rapid development in recent years, driven by advances in high-resolution computer vision models, such as Segment Anything Model and Grounding DINO [19-21]. The availability of high-resolution RGB weed datasets, including PhenoBench, DeepWeeds, and WeedsNet, has enabled the direct application of general-purpose computer vision models to weed detection tasks, as these datasets closely resemble standard high-definition RGB imagery in both spatial and spectral resolution [22-24]. This has significantly accelerated progress in ground-based weed detection. Notably, the recently proposed WeedsNet, which aims to provide a global foundation model covering over 1600 weed species using high-resolution imagery. However, the dataset and pretrained models have not yet been publicly released [24].

In contrast, UAV-based weed detection poses additional challenges [25]. UAV imagery is commonly captured at coarser spatial resolutions, and as GSD increases, models pretrained on ground-based imagery require more extensive retraining to transfer effectively. These difficulties are further intensified when transitioning from RGB to multispectral or hyperspectral data, since the inclusion of additional spectral channels increases the need for channel-specific retraining to fully leverage the discriminative power of the model [25].

Weed detection has emerged as one of the fastest-growing applications in UAV-based agricultural monitoring, driven by rapid advances in sensing technologies and deep learning. Recent reviews indicate a strong upward trend, with deep-learning-based approaches increasing from approximately 1% of weed-detection studies in 2017 to nearly 25% by 2023–2024 [25]. DJI platforms are the most commonly used in current research, accounting for around 70% of reported UAV systems, with Phantom and Mavic models appearing most frequently [25]. Although a variety of sensors have been explored, integrated RGB cameras remain the most prevalent due to their accessibility and cost-effectiveness, while multispectral sensors, such as the Sequoia and MicaSense RedEdge, are increasingly adopted for their ability to enhance plant discrimination and assess crop health [25]. Across UAV-based weed detection studies, spatial resolutions typically range from mm to several cm, with resolutions of 1–2 cm per pixel often sufficient for individual weed detection [25]. Most approaches focus on semantic segmentation using deep learning, predominantly relying on convolutional neural networks such as U-Net and its variants, with transformer-based models only beginning to emerge [18, 25, 26]. Together, these trends highlight both the growing importance of UAV-based weed detection in precision agriculture and the need for robust, transferable models that can generalize across sensors, resolutions, and crop systems [25].

## III. Data & Methods

### A. Public Dataset

The newly released foundation dataset msuav500K includes high-resolution RGB and multispectral imagery collected using UAV-mounted sensors. Although the dataset contains data from several RGB and multispectral sensors, we only used the imagery captured by the multispectral sensors. These included the Parrot Sequoia, DJI Mavic 3 M, DJI Phantom 4 Multispectral, MicaSense Altum (PT), and MicaSense RedEdge. Depending on the sensor, the multispectral data consisted of either four or five spectral bands [2].

The msuav500k dataset is already available in a pre-processed state, meaning all multispectral imagery has been radiometrically corrected and converted to reflectance, harmonized across sensors, and min–max scaled per channel to 8-bit. This standardized 8-bit format ensures efficient data compression and is particularly beneficial for SSL approaches [9, 10].

In this study, we further restricted our analysis to the spectral bands that are available across all sensors, which are Green, Red, Red-edge, and NIR, as these bands are widely used for vegetation monitoring and agricultural mapping [5]. This selection also ensured that the resulting pretrained models can be applied to imagery from any multispectral sensor commonly used in fields like precision agriculture.

Multispectral chips containing only no-data values were removed (in total 1032), as retaining them would have required additional band masking during training to exclude them from mean-std normalization, which would have increased computational cost for little expected benefit. After this filtering, the multispectral dataset consisted of 85925 image chips with a size of 512×512 pixels.

### B. Data collection

Data collection was performed using a custom-built hexacopter equipped with a MicaSense RedEdge 3 camera developed by the Agrotechnology Research Group at the University of Helsinki [27]. The camera captured five spectral bands, specifically Blue (475 nm), Green (560 nm), Red (668 nm), Red Edge (717 nm), and NIR (840 nm) [27].

Imagery was acquired over a 5-year period from 2020–2024 on agricultural fields of the Viikki Research Farm of the University of Helsinki exclusively during the summer months (May to September). Two flight altitudes were used (50 m and 10 m). Flight plans were designed with 80% front and side overlap to ensure comprehensive coverage. The dataset included multiple crop types, such as barley, faba bean, oat, rapeseed, and silage grass.

The 10-m flights captured non-overlapping data, while the 50-m flights included repeated coverage of the same fields at different times. This strategy resulted in a highly diverse visual dataset representing a range of crop growth stages and conditions.

### C. Data preprocessing

The data were calibrated using reflectance target images captured directly after each flight campaign, mosaiced, and used to compute spectral indices for the Blue, Green, Red, Red Edge, and NIR bands using PIX4D software [28]. To reduce noise, all images were clipped and rescaled to a 0–1 range. The dataset was then divided into 512×512 image chips for self-supervised training, with each chip containing the five spectral channels (Blue, Green, Red, Red Edge, and NIR). While all five spectral channels were captured, the Blue band was excluded to improve cross-sensor compatibility with msuav500K and general multispectral cameras, as many sensors only use four channels (Green, Red, Red Edge, NIR).

From the generated image chips, image chips containing no-data values were excluded, resulting in 5175 image chips from the 50-m flights and 1155 chips from the 10-m flights. Preprocessed example chips of the self-collected dataset are visualized as false-color infrared (Figure 2).

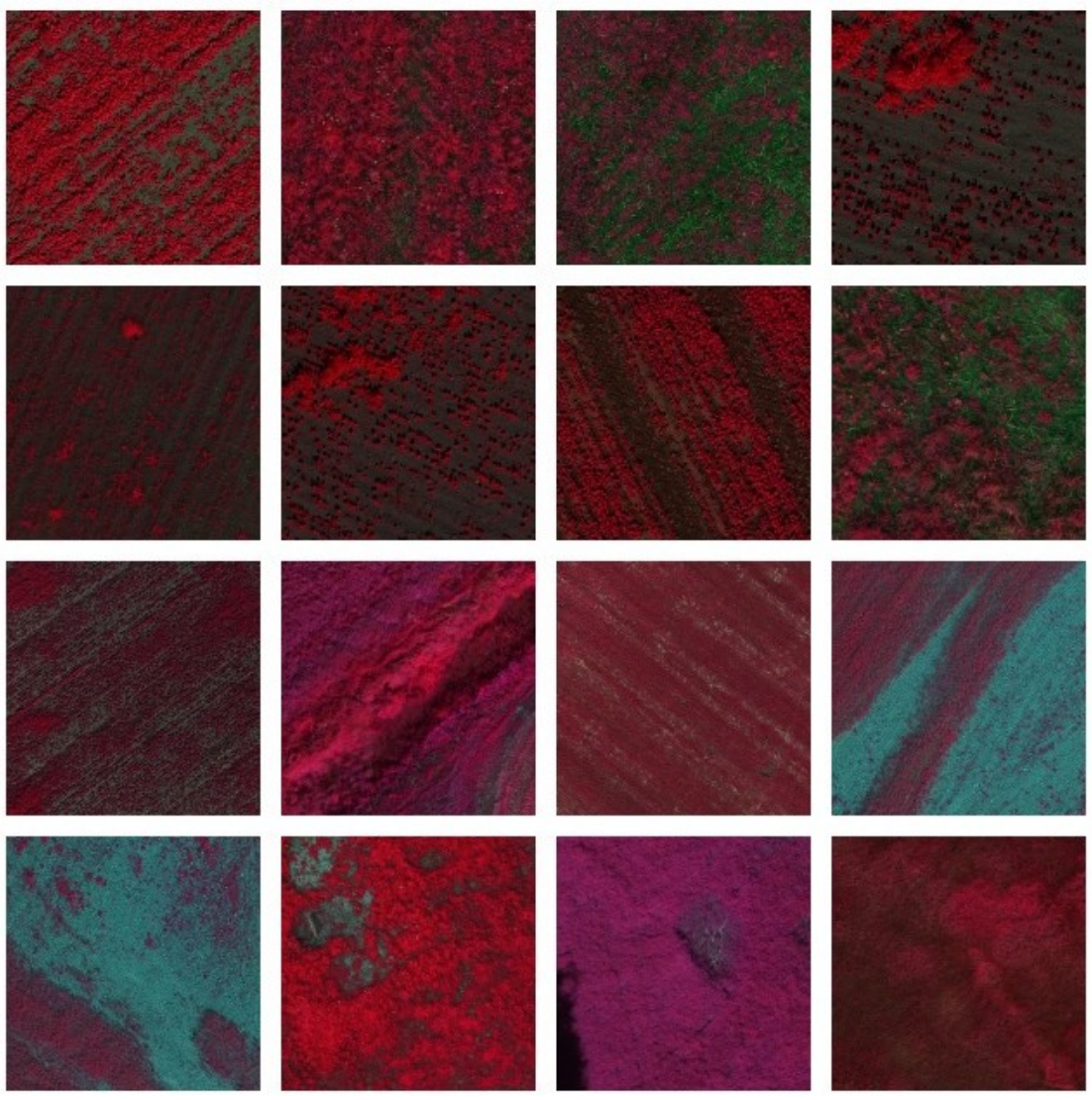

**Figure 2. Image chips of the dataset visualized as false-color infrared. The first two rows show images from data collection using 10-m altitude and last two rows from 50-m altitude.**

The 1155 image chips from the 10-m flights are publicly released alongside this publication to expand the available high-resolution UAV multispectral datasets on the platform Zenodo as four- and five-channel versions [6]. The 50-m chips are currently reserved for ongoing research but may be released in the future. Importantly, although the 50-m data included repeated imagery of the same fields under different growth conditions, these image chips constituted only approximately 5–6% of the combined dataset. Therefore, when integrated with msuav500K, it is highly unlikely that the model will overfit spatial patterns, especially given their significant visual variations across the growing season. All image chips were combined into a single dataset for training. To harmonize with msuav500K, each image chip was converted to 8-bit using a per-channel min–max scaling, consistent with the 8-bit conversion approach used in msuav500K [2]. Previous studies have shown that this scaling provides effective data compression, which reduces computational cost and memory requirements for large-scale pretraining while preserving sufficient spectral detail [9, 10, 27]. In Table I, a summary of the preprocessed image chips of the self-collected dataset is given, with more details in supplementary materials Table A1.

TABLE I

Summary of the self-collected dataset

| Crops | Altitude | GSD | Number of flight campaigns | Years | Number of Image chips with size 512x512 |
|---|---|---|---|---|---|
| Oat (*Avena sativa* L.), barley (*Hordeum vulgare* L.), faba bean (*Vicia faba* L.), rapeseed (*Brassica napus* L.) | 10 m | 0.7 cm | 7 | 2021, 2023 | 1155 |
| Barley (*Hordeum vulgare* L.), faba bean (*Vicia faba* L.), oat (*Avena sativa* L.), rapeseed (*Brassica napus* L.), and silage grass (mostly timothy, *Phleum pratense* L.) | 50 m | 3.3 cm | 51 | 2021, 2022, 2023, 2024 | 5175 |

### D. Overview of Combined Dataset

The final preprocessed dataset for self-supervised training was created by combining msuav500K with self-collected data, resulting in 92320 image chips, of which 95% (87 704) were used for training and 5% (4616) for validation.

Although validation is often omitted in self-supervised pretraining, it was included here due to the relatively small dataset size, allowing monitoring of potential overfitting. Sensors were sampled in a stratified manner to ensure consistent representation across datasets, with the self-collected data treated as a single sensor. The dataset included DJI Mavic 3M, DJI Phantom 4M, MicaSense RedEdge-M, MicaSense Altum, MicaSense RedEdge-MX, Parrot Sequoia, MicaSense Altum-PT, DJI Zenmuse X3 NIR, and MicaSense RedEdge 3. Only the Green, Red, Red-Edge, and NIR bands were retained; any Blue channels were discarded.The dataset spanned multiple geographies (Italy, UK, Germany, Sri Lanka, Portugal, Mexico, Côte d'Ivoire, Spain, Brazil, Greece, Serbia, Slovenia, Belgium, USA, Poland, and Finland) and diverse land uses, including agriculture, forests, mining areas, and nature reserves. While most regions were present in the original msuav500k datasets, our extensions added low-value crops at various growth stages, particularly in Finland using the MicaSense RedEdge 3. Table A2 in the annex summarizes mean and standard deviation (std) per spectral band across datasets and is provided to show compatibility and provide insight into the data distribution.

### E. Downstream task A: WeedMap (MicaSense RedEdge-M).

For downstream task A evaluation, we used the publicly available WeedMap dataset introduced by Sa et al. [5]. The dataset was collected in agricultural fields in Germany and Switzerland at a flight altitude of 10 m using a UAV-mounted camera, resulting in a GSD of approximately 1 cm. For the downstream task A, we used the MicaSense RedEdge-M data collected in Germany. The dataset covers a sugar beet field in Rheinbach, where crops were planted on 18 August 2017 and data were collected 1 month later on 18 September 2017 [5]. Crops were sown in rows with 50-cm spacing between rows and 20-cm spacing within rows.

At the time of acquisition, sugar beet plants had heights of approximately 15–20 cm, while weeds measured 5–10 cm. Minimal mechanical weed control was applied, resulting in a high weed density in the field [5]. The original semantic classes, Background (0), Crop (1), and Weeds (2), were remapped to Background (1), Crop (2), and Weeds (3), since class 0 is reserved for no-data or unclassified pixels in the SpecDeepMap API and a modified implementation of the SpecDeepMap API will be used for the downstream tasks experiments [29]. The multispectral images were min–max scaled to 8-bit and split into 224×224 image chips with the channel Green, Red, Red-Edge, and NIR. Only full-label chips were included in the dataset and chips that were not only representing the background (0) class. Further noisy and corrupted label chips present in the dataset were filtered, resulting in 739 image chips. Furthermore, the predefined spatial independent test orthomosaic of the dataset was split separately into 224×224 with same spectral bands, resulting in 280 image chips for testing.

To evaluate the benefits of pretraining under limited-data regimes, the pretrained encoders were fine-tuned using varying fractions of the downstream training data (5%, 10%, 25%, 50%, 75%, and 100%), drawn from the 90% portion of the dataset allocated for training, while the remaining 10% was reserved for validation. To ensure comparable class distributions across all data subsets, we employed a one-dimensional Wasserstein-distance-based optimization using label distributions [30]. For this purpose, we adapted and extended an approach originally developed for the Dataset Maker algorithm of the EnMAP-Box application SpecDeepMap [29].

In the first stage, we took all 739 available image chips and summarized their label histograms and divided them by their total label counts and converted them into discrete probability distributions, which were stored in a matrix $H \in \mathbb{R}^{N \times C}$, where $C$ signifies the number of classes and $N$ the number of images. We then performed 5000 random permutations in which the data were split into 90% training and 10% validation subsets. For each permutation, we computed the one-dimensional Wasserstein distance ($W_1$) between the class distributions of the two subsets [29-31].

$$W_1(H_{\text{train}}, H_{\text{validation}}) = \sum_{c=1}^{C} \mid F_{\text{train}}(c) - F_{\text{validation}}(c) \mid$$

Where C denotes the total number of classes and $F_{\text{train}}$ , $F_{\text{validation}}$ are the cumulative distribution function of the subset histograms, representing the discrete class probability distribution of each subset [29-31].

$$F_{\text{train}}(c) = \sum_{i=1}^{c} H_{\text{train}}[i]$$

$$F_{\text{validation}}(c) = \sum_{i=1}^{c} H_{\text{validation}}[i]$$

The training-validation subsets of all 5000 permutations with the minimum Wasserstein distance was selected, as it exhibited the most similar class distributions.

$$p^* = \arg\min_{p} \quad W_1(p)$$

In the second stage, the selected training dataset was cumulatively subdivided into increasing fractions of 5%, 10%, 25%, 50%, 75%, and 100%. For each fraction, a similar Wasserstein-based optimization was applied. Starting with the 5% split, we randomly sampled candidate subsets again using 5000 permutations and computed the one-dimensional Wasserstein distance between the selected subset and the remaining training data at each permutation [29-31]. The subset with the minimum Wasserstein distance among all permutations was chosen. This procedure was repeated iteratively for larger fractions, with the constraint that each larger split fully contained all smaller splits; previously selected samples were therefore excluded from subsequent permutation pools.

This strategy ensured consistent class balance across all data fractions and enabled a fair evaluation of performance gains from pretraining, since each smaller dataset was entirely contained within the larger ones. The test field data were excluded from the Wasserstein distance optimization to preserve spatial independence, and all 280 test data image chips were used for the model assessments.

### F. Downstream task B: WeedMap (Sequoia)

To evaluate cross geographical and cross-sensor generalization, we further assessed the best performing self-supervised pretrained model from downstream task A on a second downstream task, referred to as downstream task B. This task also used data from the publicly available WeedMap dataset introduced by Sa et al. but relied on imagery acquired with a different sensor and at a different geographical location [5].

The WeedMap Sequoia data were collected over two agricultural fields in Switzerland growing sugar beet (*Beta vulgaris*) during two UAV flight campaigns conducted on 5 April 2017 and 17 March 2017. At the time of data acquisition, crops were sown in rows with a spacing of 50 cm between rows and approximately 18-cm spacing within rows [5]. The imagery was captured using a Sequoia multispectral camera, providing the spectral bands Green, Red, Red Edge, and NIR.

Data preprocessing and balanced sub-dataset construction for downstream task B followed the same procedure as described for downstream task A. Specifically, the multispectral images were min–max scaled to 8-bit and divided into 224×224 pixel image chips using the Green, Red, Red Edge, and NIR channels. Only fully labeled image chips were retained, excluding chips that contained only background pixels. Furthermore, the original semantic classes labels, namely Background (0), Crop (1), and Weeds (2), were remapped to Background (1), Crop (2), and Weeds (3), as in downstream task A.

Training and validation splits were generated using the identical one-dimensional Wasserstein distance-based optimization strategy employed for downstream task A, ensuring comparable class distributions between the subsets. The dataset was first divided into 90% training and 10% validation data. The training set was then further subdivided into cumulative class balanced subsets of 5%, 10%, 25%, 50%, 75%, and 100%, with each smaller subset fully contained within all larger subsets. In total, this process resulted in up to 591 image chips for training and 66 image chips for validation. The test dataset consisted of 62 spatially independent image chips, which were excluded from the Wasserstein distance optimization and were used exclusively for model evaluation.

### *G. Experimental set up*

#### I. Self-supervised Training

For SSL, MoCo-v3 and MAE were used [3, 12]. MoCo-v3 incorporates extensive data augmentation techniques, such as scale, brightness, contrast, grayscale, and smoothing rotation. However, we followed the common practice in multispectral pretraining and excluded color jitter augmentation. These augmentations encourage the encoder models to learn more generalizable features throughout the training. During pretraining, we limited brightness and contrast augmentation in our training to a range of 0.85–1.15 with a 30% random chance, rather than the commonly applied 0.6–1.4 with an 80% chance, to reduce spectral distortions [12]. We also omitted the commonly used grayscale and solarize augmentations [12]. Furthermore, to ensure correct normalization, the data were normalized after the augmentation with mean and std per channel based on the training dataset.

We evaluated the encoder architecture Swin Transformer, namely Swin-s3_tiny_224, with MoCo-v3 [32]. We pretrained the Swin Transformer in a self-supervised manner using MoCo-v3 with an effective batch size of 252 by utilizing gradient accumulation three for 200 epochs on a single GPU with initial learning rate of 0.0001, which was derived from the original learning rate for MoCo-v3 pretraining experiments. With the 200 epochs, we followed a training duration that is commonly applied for MoCo training length and selected the model that will be used for downstream experiments before self-supervised loss convergence stops; this was at epoch 79 in our pretraining.

Using the MAE framework, we evaluated the following two ViT architectures: ViT Small (Patch 16) and ViT Base (Patch 16)[3, 33]. MAE is a widely adopted SSL approach that randomly masks input patches and trains the model to reconstruct the missing content from the visible patches[3]. To achieve an effective batch size of 256, we employed gradient accumulation with four steps (4×64). ViT Small (Patch 16) and ViT Base (Patch 16) were pretrained in a self-supervised manner using MAE for 400 epochs on a single GPU. In contrast to MoCo training schedules, MAE trainings were generally longer and selected the model that will be used for downstream experiments before self-supervised loss convergence stops, which was at epoch 400 for ViT Small (Patch 16) and ViT Base (Patch 16). In original MAE experiments, it was shown that MAE benefits from training runs twice as long compared with MoCo-v3 that saturates earlier, meaning the self-supervised loss convergence stops [3]. For both ViTs we trained MAE an initial learning rate of 0.03 scaled according to batch size and applied cosine learning-rate decay over the full training schedule and a masking ratio of 0.75.

During pretraining, these image chips were randomly cropped to 224×224 pixels, with scale 0.2–1, a commonly used scale resizing in SSL pretraining to support learned feature generalization, also for high-resolution datasets [9, 10, 34].

#### II. Downstream Tasks

The pretrained models were fine-tuned on each training split of the downstream task for 50 epochs. The model with the highest validation Intersection over Union (IoU) score of each training run was evaluated on the test dataset [35]. Fine-tuning on each training data split was conducted using a batch size of 8 and an initial learning rate of 0.0003. ViTs were implemented within the Dense-Prediction Transformer (DPT) architecture and Swin Transformers with U-Net architecture [32, 36, 37].

During fine-tuning, the pretrained encoders were evaluated frozen and unfrozen (full-fine tune setting) and their performance was compared against the following baselines: models with unfrozen and frozen encoders initialized with ImageNet weights and models with unfrozen encoders trained from randomly initialized weights. In case of Swin Transformer encoder, this model was also compared against an unfrozen and frozen Swin Transformer trained by Doornbos et al. 2025 on a prerelease version of msuav500k, previously called msuav100k [7]. Standard data augmentation and normalization were applied, using per-channel mean and std, and balanced training was achieved through class weighting based on inverse class frequency [38].

For downstream task evaluation, we employed a custom adaptation of the SpecDeepMap API, extended to incorporate MAE and MoCo-v3 model encoders that were pretrained as part of this study, within an unreleased API version together with the DPT architecture [29].

## IV. Results

### A. Downstream Task A

We compare performance results obtained under different weight-initialization strategies here. The size of the training dataset is reported as a percentage, where 100% corresponds to 739 images. The validation split of 79 images was kept constant across all experiments. All models were evaluated on a spatially independent test dataset of 280 image chips on mean IoU [35]. Each model was assessed using several weight configurations with unfrozen and frozen encoders [36, 37].

For the SSL setting, ViTs were pretrained using MAEs and ViT Small was also pretrained with MoCo-v3. Our Swin Transformer was pretrained only with MoCo-v3, as standard MAE can not be implemented directly with hierarchical feature extraction of Swin Transformers [3, 12, 33, 39]. The results of the experiments in Figures 3–5 show the mean IoU on the test dataset of each model according to which weight initialization and training data volume were used and whether the encoder weights were frozen or unfrozen, meaning if partial or full fine-tuning of the model was applied.

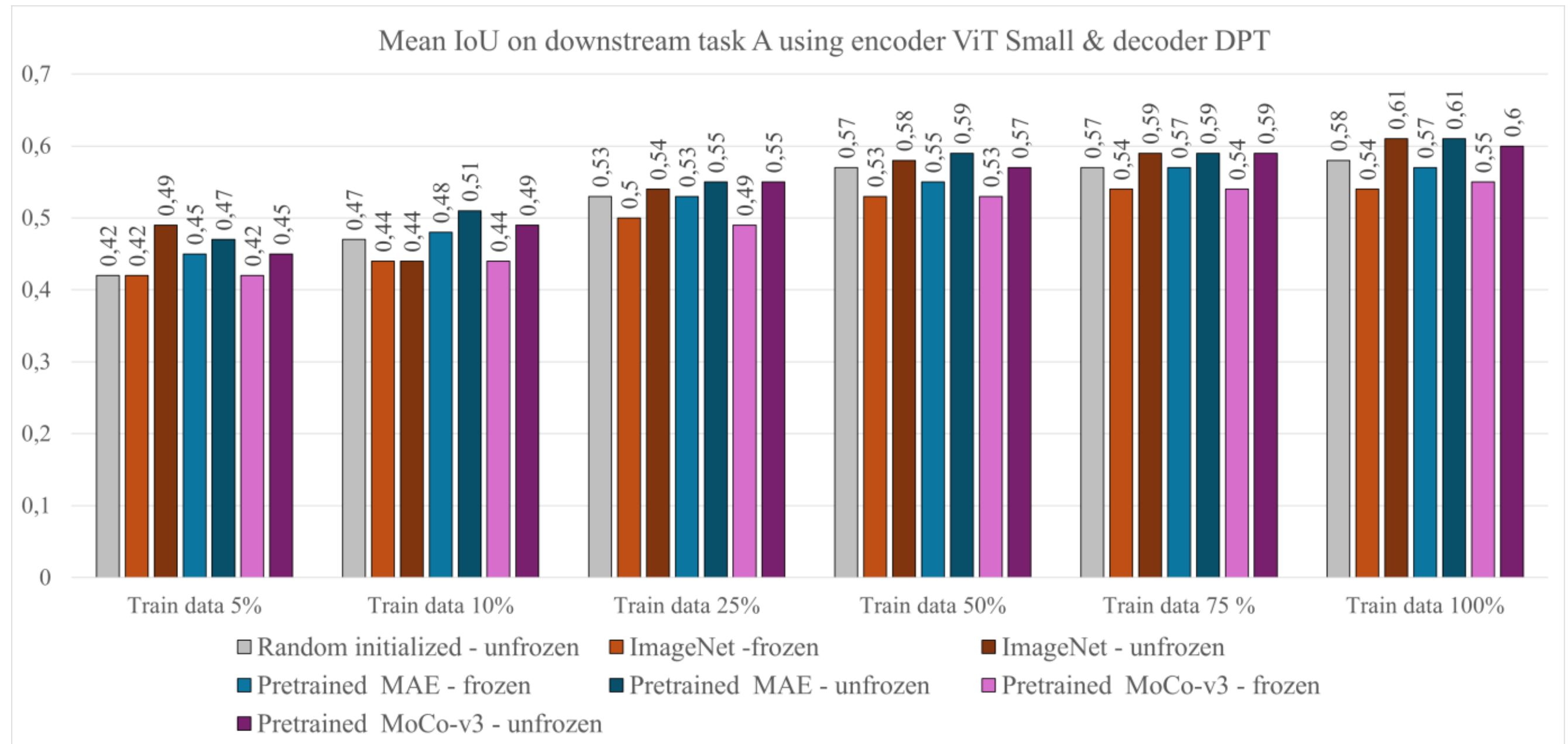

**Figure 3.** **Mean IoU on test dataset using different training data volumes and encoder ViT Small (Patch16) and decoder DPT.**

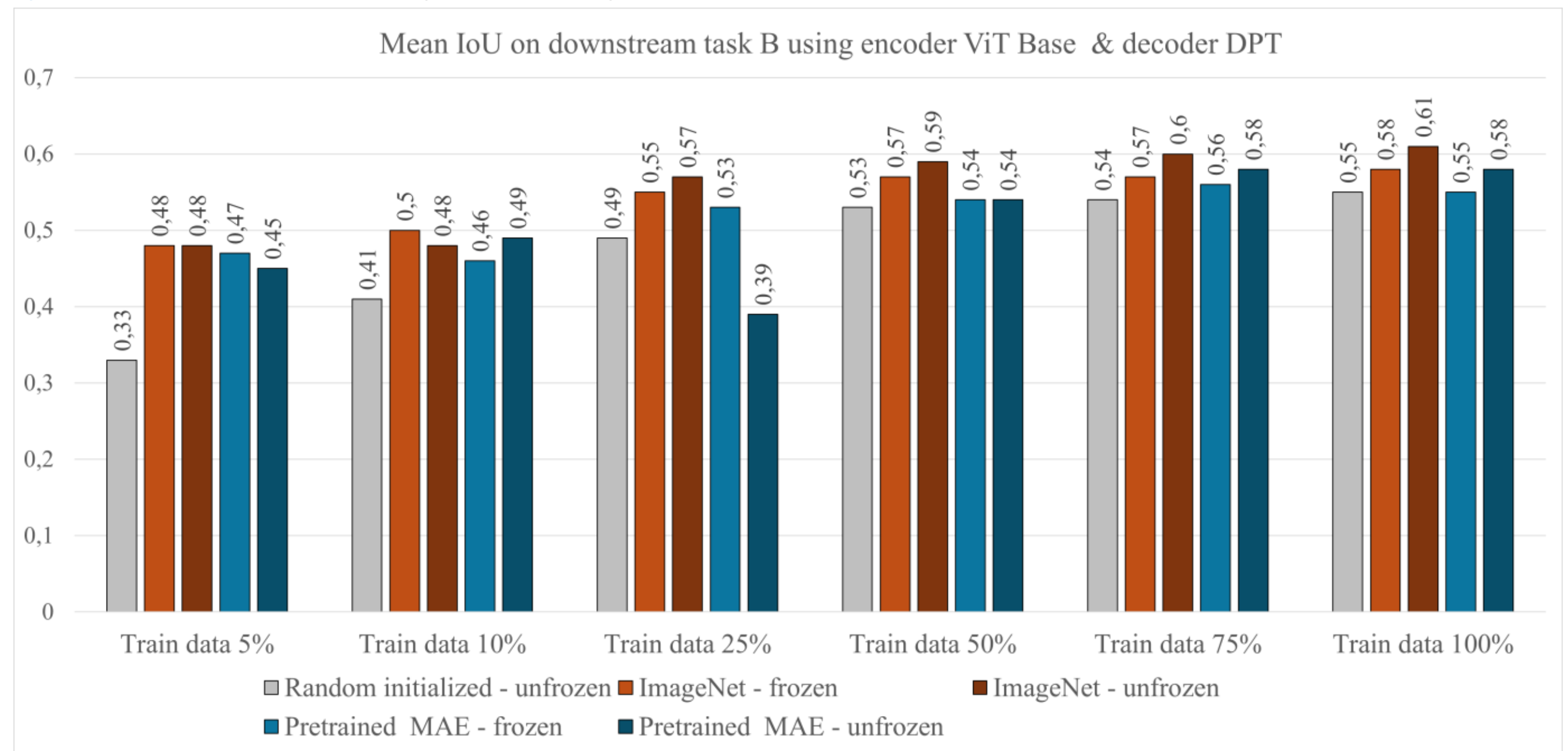

**Figure 4.** **Mean IoU on test dataset using different training data volumes and encoder ViT Base (Patch16) and decoder DPT.**

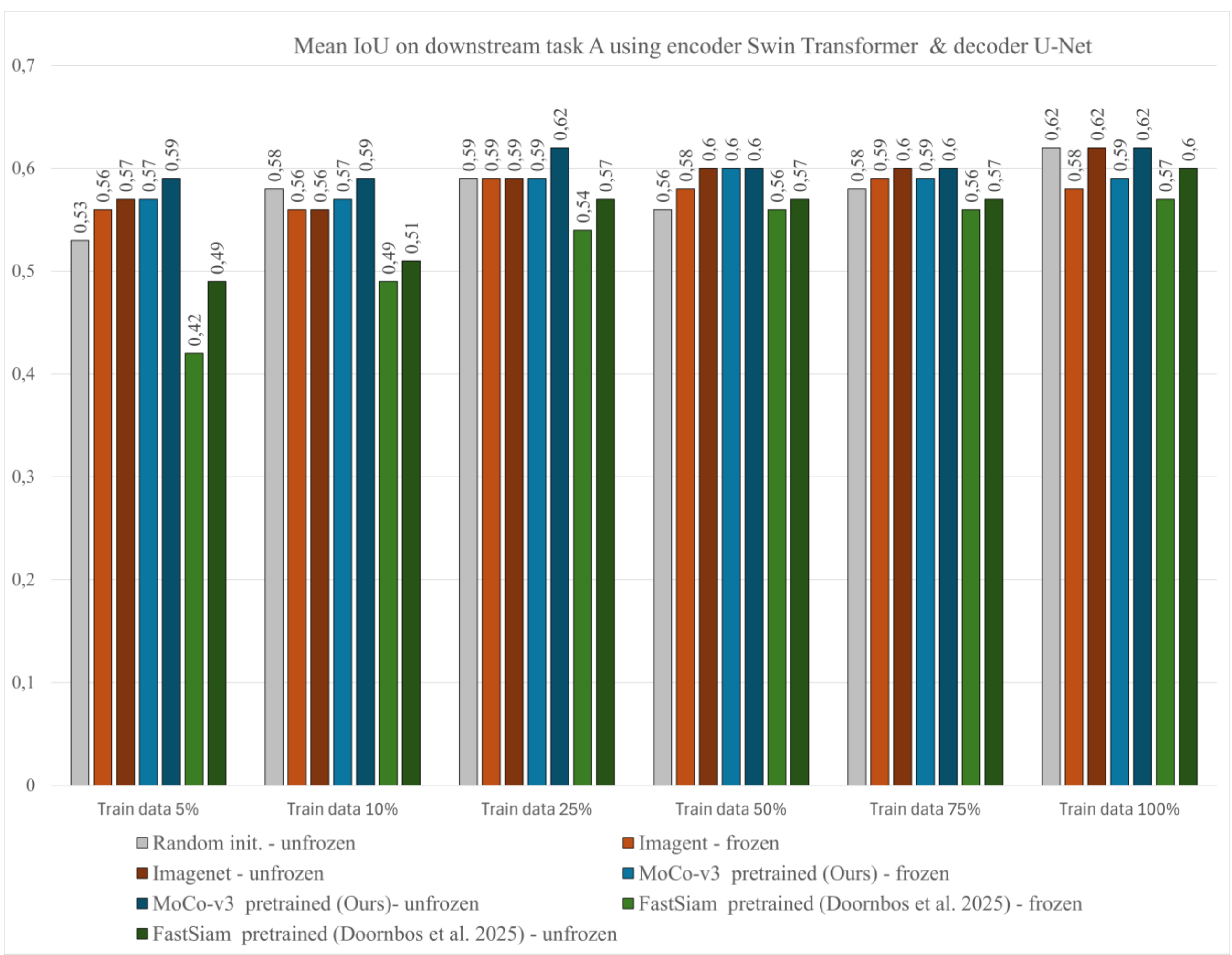

**Figure 5. Mean IoU on test dataset using different training data volumes and encoder Swin Transformer and decoder U-Net.**

Pretrained MAE ViT-Small (patch16) demonstrated particularly strong performance in low-data regimes on downstream Task A (Figure 3). In frozen-weight settings, the MAE-pretrained ViT-Small on msuav500k+N consistently outperformed both frozen ImageNet-pretrained weights and frozen MoCo-v3 pretrained weights. Under full fine-tuning, ImageNet-pretrained weights achieved the best performance at the 5% training data split. However, for 10–100% training data splits, the ViT-Small MAE-pretrained weights surpassed ImageNet weights. Across all splits, MAE-pretrained weights also outperformed MoCo-v3 pretrained weights for the ViT-Small patch16 encoder, with unfrozen MAE weights consistently achieving equal or higher IoU on 25% and 75% splits.

For ViT-Base, pretrained MAE weights mostly outperformed scratch-trained models, despite an on-training data volume of 25%, while ImageNet-pretrained ViT-Base attained the highest absolute performance (Figure 4). In extremely low-data settings, the scratch-trained model achieved an IoU of 0.33 compared with 0.45–0.48 for pretrained models, highlighting the substantial benefit of pretraining in these scenarios. Only when >50% training data volume was used along with random initialization and full encoder unfreezing did the ViT-Base achieve comparable performance.

During early training stages and under limited data conditions, VIT Base outperformed the smaller ViT Small in frozen-weight settings (Figures 3 and 4). As the amount of labeled data increased, this advantage diminished. However, under full fine-tuning, ViT-Small can outperform ViT-Base on 5% and 10% splits and achieved comparable performance from 50–100% of training data (Figures 3 and 4).

For the Swin Transformer, the variant pretrained with MoCo-v3 on msuav500k+N and fully unfrozen outperformed all other weight configurations on the 5–25% data splits and remained competitive across the 50–100% training data volume (Figure 5). Our pretrained Swin Transformer, when unfrozen, outperformed both frozen and unfrozen Swin Transformer variants pretrained by Doornbos et al. across all splits [7]. The improvements were most pronounced in low-data settings, with frozen Swin Transformer achieving 15 IoU higher than the Doornbos et al variant on the 5% split and 8 IoU higher on the 10% split, while unfrozen Swin achieved 10 and 8 IoU higher on 5% and 10% splits, respectively (Figure 5). From 25–100% splits, unfrozen Swin Transformer consistently outperformed the Doornbos pretrained variant by 2–5 IoU. Compared with ViT-Small and ViT-Base, Swin Transformer achieved approximately 10 IoU higher at 5%, 5–8 IoU higher on 10–25%, and 1–2 IoU higher on 50–100% splits (Figures 3–5). The combination of a U-Net decoder and hierarchical window attention mechanism, which allows very small patch sizes of 4 at the first level, likely contributes to the strong performance of Swin Transformers in fine-grained segmentation. Whereas the implemented ViTs used a patch size of 16, as a patch size of 4 with the ViT architectures and DPT was computationally prohibitive, as reducing the patch size to 4 would increases the number of tokens by 16-fold and the self-attention cost with the ViT encoders by approximately 256-fold compared to the patch 16 configuration.

### *B. Downstream Task B*

Here, we further evaluate the best performing pretrained encoder from downstream task A on a second downstream task B to evaluate cross-sensor and cross-geographical performance. Again, we compare performance results obtained under different weight initialization strategies and frozen and unfrozen model weight settings. The best performing encoder in downstream task A was our MoCo-v3 pretrained Swin Transformer paired with U-Net decoder. The size of the training dataset is reported as a percentage, where 100% corresponds to 591 image chips for training. The validation split of 66 images was kept constant across all experiments. All models were evaluated on a spatially independent test dataset of 62 image chips on mean IoU for each model-weight initialization and training data volume used.

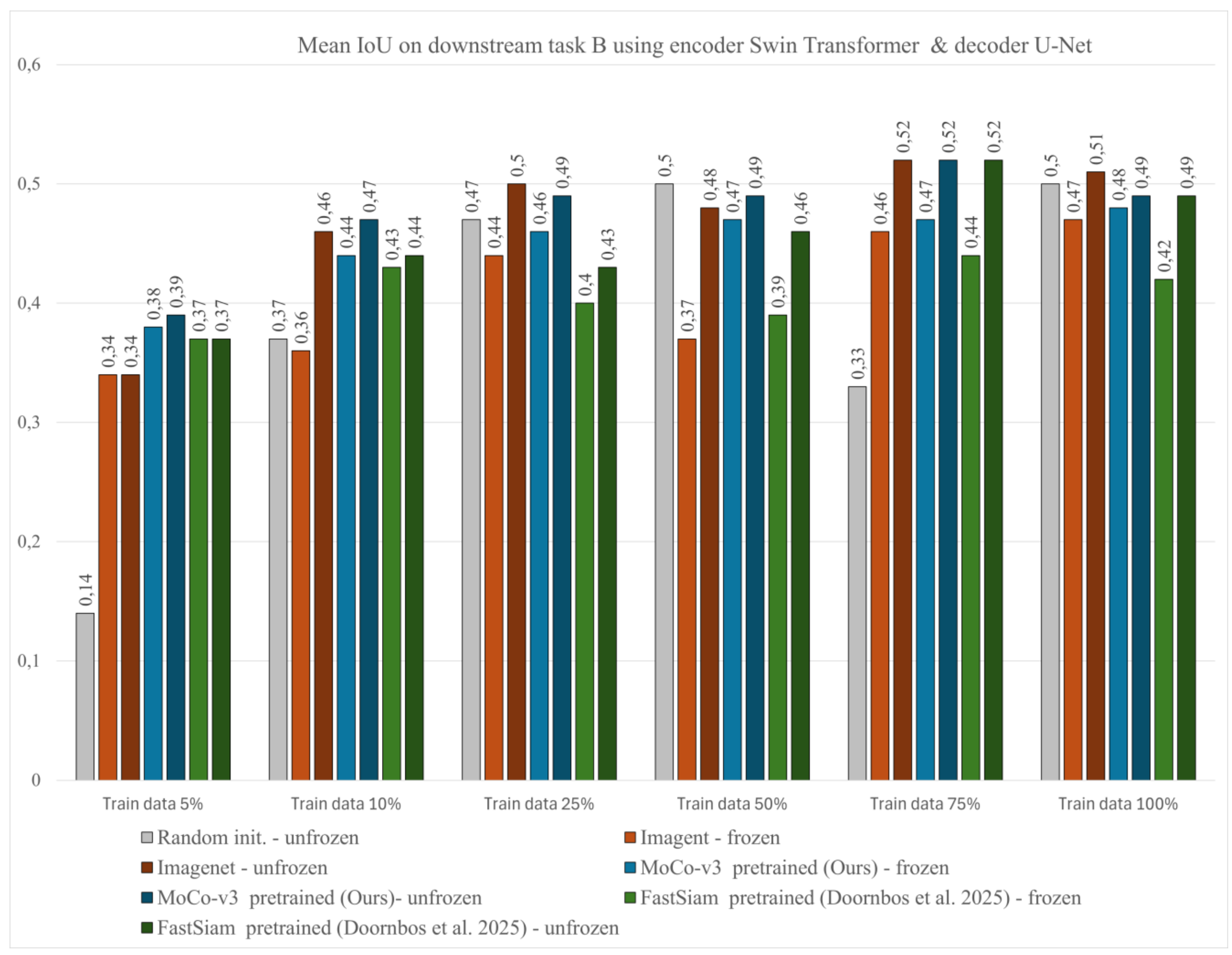

**Figure 6.** **Mean IoU on test dataset using different training data volumes and encoder Swin Transformer and decoder U-Net.**

On downstream Task B, which was the data collection from Switzerland using the Sequoia sensor of the WeedMap dataset, our pretrained Swin Transformer achieved the best performance on 5–10% splits (Figure 6). On the 5% split, the unfrozen pretrained encoder outperformed the worst-performing scratch-trained model by 25 IoU and by 11 IoU on the 10% split relative to frozen ImageNet initialization. Overall, it achieved the best performance with IoU of 0.52 on the dataset at the 75% training data used, matching the ImageNet unfrozen model that also achieved 0.52 IoU. Across 25–100% splits, this was 1–2 IoU lower than the best-performing model in each range. Our pretrained Swin Transformer also outperformed the pretrained variant by Doornbos et al. on 5–50% data training volume and performed equivalently on 75–100% [7].

## V. Discussion:

All research hypotheses are discussed here.

**Hypotheses (i):** *Self-supervised pretraining on the combined msuav500k+N dataset improves semantic segmentation performance on downstream Task A, particularly in low-data regimes, compared with (a) sensor- and geography-specific models trained end-to-end from scratch and (b) models initialized with ImageNet-pretrained weights. Evaluation is conducted on the WeedMap dataset collected in Germany using the RedEdge-M sensor.*

The results support this hypothesis. In case of the Swin Transformer architectures, self-supervised pretrained models consistently outperformed models trained from scratch and those initialized with ImageNet weights when only a small fraction of the training data was available in downstream task A and B. In case of the ViT models, both self-supervised pretrained variants on msuavk500+N surpassed scratch end-to-end trained models from random weight initialization, although they did not surpass the full fine-tuned ImageNet weight pretrained model in the low-data regime, meaning 5–10% training data. On Task A, MAE-pretrained ViT-Small achieved substantially higher IoU in frozen weight settings than both frozen ImageNet and MoCo-v3 weights, particularly on the 5% and 10% splits, while unfrozen MAE weights maintained strong performance at intermediate splits, such as 25% and 75%. Similarly, the MoCo-v3 pretrained Swin Transformer in the unfrozen configuration outperformed all other weight settings on 5–25% splits of downstream task A and remained competitive up to 100% of the training data. These results highlight the effectiveness of self-supervised pretraining for extracting robust features from large multispectral datasets, which can be transferred to downstream tasks even when labeled data are extremely limited. The performance gap between scratch-trained models and pretrained models, which can exceed 0.6–0.25 IoU at the 5% split, underscores the substantial benefit of pretraining in low-data regimes. However, all experiments showed that with increase of training-data volume, these competitive performance gains diminish.

**Hypothesis (ii):** *The best-performing self-supervised pretrained model trained on msuav500K+N demonstrates cross-geographical and cross-sensor generalization, achieving competitive performance on downstream Task B using the WeedMap dataset collected in Switzerland with the Sequoia sensor.*

This hypothesis is also supported by the results. On Task B, which involves a sensor and geographic domain distinct from downstream task A, the pretrained Swin Transformer achieved the highest performance on 5–10% splits, outperforming scratch-trained models by 25 IoU at 5% and 11 IoU at 10% and achieving still the best overall IoU on the dataset of 0.52 at the 75% split, matching the performance of ImageNet-unfrozen weights. These findings indicate that the representations learned during self-supervised pretraining on the combined msuav500K+N dataset were not only effective for one sensor or geography but also generalized well across two different geographical contexts (Germany and Switzerland) and across sensors Rededge-M and Sequoia. The consistent superiority of unfrozen variants in low-data splits suggests that fine-tuning is critical to adapting pretrained features to the new domains while retaining the robustness conferred by the pretraining.

**Hypothesis (iii):** *The best-performing self-supervised pretrained model trained on msuav500K+N outperforms the Swin Transformer trained on msuav500K by Doornbos et al. (2025), which was based on a pre-release multispectral-only version of the dataset.*

The results provide support for this hypothesis. Across low- and mid-data splits, our pretrained Swin Transformer consistently outperformed the Doornbos et al. Swin Transformer variant, with differences reaching 8–10 IoU in 5–10% splits, and maintained an advantage of 2–5 IoU from 25–100% splits (Figures 5 and 6). These improvements likely resulted from longer pretraining schedules and further could be influenced by the extended pretraining dataset and the transformer-optimized SSL approach (MoCo-v3) compared with the self-supervised method FastSiam, which was utilized in pretraining by Doornbos [7]. However, it should also be noted that the Swin Transformers are not exactly the same variants. While we both used the tiny Swin Transformer variant, we utilized in our self-supervised pretraining the third Swin transformer version, while Doornbos et al. pretrained the first Swin transformer version [32].

Overall, the findings related to all three hypotheses show that self-supervised pretraining on the large, multispectral dataset msuav500k+N improves performance of the models especially in low-data regimes and could also demonstrate cross-domain generalization in case of the Swin Transformer, with full fine-tuning further improving results. Furthermore, the results showed that model type influences the segmentation performance results and that the Swin-U-Net combination outperforms ViT Small or ViT Base combined with DPT, especially in the low training data regime of 5–10%.

## VI. Limitations

Although self-supervised pretraining on the combined msuav500K+N dataset provides clear improvements in downstream semantic segmentation in low-data regimes, several factors still influence model performance and generalization. One consistent observation is that ImageNet-pretrained weights remain highly competitive. This reflects the high-quality, fine-grained spatial representations learned from the extremely large and diverse ImageNet dataset, which are particularly well-suited for tasks such as early-stage weed detection, where vegetation is sparsely distributed in an agricultural field and recognition of fine spatial patterns is important. Furthermore, as the domain gap between ImageNet imagery and high-resolution four-band UAV imagery is not that large compared with e.g. Landsat imagery, as they are substantially closer to standard RGB imagery in both fine-grained patterns and visual appearance. Another consideration is the heterogeneity of the pretraining dataset. Although msuav500K+N spans multiple sensors and geographic regions, differences in spectral-band centers, bandwidths, and flight conditions introduce variability that can complicate representation learning. More critically, the msuav500K dataset covers an extreme GSD range from 0.1–25 cm, corresponding to a scale variation of roughly 250 times [2]. Such large-scale discrepancies limit the transferability of representations learned by MoCo and MAE and make the domain gap between pretrained and end-to-end scratch-trained models closer [40].
Finally, although our SSL pretraining demonstrated gains across architectures, tasks, and sensors, performance differences across data splits indicate that the benefits of SSL are more pronounced in extremely low-data settings. In higher-data regimes, the performance advantage over ImageNet-pretrained weights or scratch end-to-end diminishes, suggesting that further improvements may require larger-scale or more scale-homogeneous pretraining datasets, extended training schedules, or scale-aware methods that explicitly model spatial frequency variations. Future work could explore several directions to improve SSL pretraining. One option is that pretraining could be restricted to narrower GSD ranges to reduce absolute scale variability and better match models to specific target resolutions. Another promising direction is the use of scale-aware methods such as Scale-MAE, which explicitly model high- and low-frequency representations [40]. However, existing studies using Scale-MAE reported only modest gains for semantic segmentation [40]. Moreover, Scale-MAE requires known GSD values for each image, while msuav500K lacks per-image GSD metadata. While the GSD for most sub-data collection is listed in the msuav500k publication, it also includes data from several flight campaigns with undocumented GSDs, making the application of Scale-MAE non-trivial. Although this was not implemented within the scope of this research, this could be an interesting approach for future studies [2, 40].

## VII. Conclusion

This study demonstrates that self-supervised pretraining on the combined msuav500K+N dataset substantially improves semantic segmentation performance for multispectral UAV imagery, particularly in low-data regimes. Across architectures and downstream tasks, pretrained models consistently outperform scratch-trained models and, in many cases, surpass ImageNet-initialized baselines when only limited labeled data are available (5–10%). On downstream Task A, MAE-pretrained ViTs and MoCo-v3-pretrained Swin Transformers showed clear advantages at 5–25% training splits, highlighting the importance of large-scale, domain-relevant pretraining when annotations are scarce. The Swin Transformer with MoCo-v3 pretraining achieved the strongest overall performance, benefiting from hierarchical feature extraction and fine-grained spatial modeling, and consistently outperformed the previously published Swin Transformer model pretrained by Doornbos (2025), especially in low-data settings [7].
Furthermore, the best-performing pretrained Swin Transformer generalized effectively across geographical regions and sensors, achieving competitive or superior results on downstream Task B in Switzerland using Sequoia sensor data. As msuav500k+N has a high scale variance from 0.1–25 cm, for future pretraining approaches we recommend that future research using this dataset or msuav500k should consider scale-aware approaches, such as Scale-MAE.
To further support UAV remote-sensing research, we publicly release 1155 high-resolution multispectral image chips (approximately 0.7 cm GSD) from seven flight campaigns covering multiple crop types. This dataset will help address the lack of openly available close-range multispectral UAV data from Nordic countries (such as Finland) and will contribute to the availability of imagery from non-high-value crops. Furthermore, the best performing encoder from our pretraining on msuav500k+N will be made available via the self-developed SpecDeepMap application, which is part of the QGIS Plugin EnMAP-Box in 2026, to make it accessible in an open-source no-code software environment for precision agriculture and other UAV research [29].

## IX. Supplementary Material

**TABLE A1.** OVERVIEW OF FLIGHT CAMPAIGN FOR THE SELF-COLLECTED DATASET

| Cultivated crop type | Altitude | GSD | Date | Number of image chips |
|---|---|---|---|---|
| Oat | 10 m | 0.7 cm | 18.06.2021 | 121 |
| Faba Bean/diverse (after harvest) | 10 m | 0.7 cm | 23.08.2021 | 177 |
| Rapeseed | 10 m | 0.7 cm | 18.06.2021 | 255 |
| Faba Bean | 10 m | 0.7 cm | 19.06.2023 | 160 |
| Faba Bean | 10 m | 0.7 cm | 15.06 2023 | 160 |
| Rapeseed | 10 m | 0.7 cm | 12.06.2023 | 147 |
| Barley | 10 m | 0.7 cm | 23.05.2023 | 135 |
| Barley | 50 m | 3.3 cm | 1.8.2020 | 126 |
| Barley | 50 m | 3.3 cm | 15.7.2020 | 124 |
| Barley | 50 m | 3.3 cm | 28.6.2020 | 135 |
| Faba bean | 50 m | 3.3 cm | 12.7.2020 | 94 |
| Faba bean | 50 m | 3.3 cm | 26.7.2020 | 89 |
| Barley | 50 m | 3.3 cm | 1.7.2021 | 118 |
| Barley | 50 m | 3.3 cm | 13.7.2021 | 121 |
| Barley | 50 m | 3.3 cm | 17.6.2021 | 125 |
| Barley | 50 m | 3.3 cm | 26.7.2021 | 115 |
| Barley | 50 m | 3.3 cm | 4.6.2021 | 124 |
| Faba bean | 50 m | 3.3 cm | 10.6.2021 | 111 |
| Faba bean | 50 m | 3.3 cm | 15.7.2021 | 115 |
| Faba bean | 50 m | 3.3 cm | 2.7.2021 | 114 |
| Faba bean | 50 m | 3.3 cm | 27.7.2021 | 110 |
| Oats | 50 m | 3.3 cm | 19.6.2021 | 87 |
| Oats | 50 m | 3.3 cm | 23.7.2021 | 98 |
| Oats | 50 m | 3.3 cm | 5.8.2021 | 102 |
| Oats | 50 m | 3.3 cm | 7.7.2021 | 93 |
| Silage grass | 50 m | 3.3 cm | 10.6.2022 | 261 |
| Silage grass | 50 m | 3.3 cm | 23.9.2022 | 260 |
| Silage grass | 50 m | 3.3 cm | 4.5.2022 | 272 |
| Barley | 50 m | 3.3 cm | 10.7.2022 | 78 |
| Barley | 50 m | 3.3 cm | 13.6.2022 | 75 |
| Barley | 50 m | 3.3 cm | 18.5.2022 | 79 |
| Barley | 50 m | 3.3 cm | 2.9.2022 | 83 |
| Barley | 50 m | 3.3 cm | 21.7.2022 | 80 |
| Barley | 50 m | 3.3 cm | 21.8.2022 | 74 |
| Barley | 50 m | 3.3 cm | 21.9.2022 | 77 |
| Barley | 50 m | 3.3 cm | 25.5.2022 | 76 |
| Barley | 50 m | 3.3 cm | 28.7.2022 | 76 |
| Barley | 50 m | 3.3 cm | 30.6.2022 | 73 |
| Barley | 50 m | 3.3 cm | 4.8.2022 | 79 |
| Barley | 50 m | 3.3 cm | 6.6.2022 | 79 |
| Rapeseed | 50 m | 3.3 cm | 3.8.2022 | 93 |
| Rapeseed | 50 m | 3.3 cm | 6.5.2022 | 114 |
| Rapeseed | 50 m | 3.3 cm | 7.9.2022 | 100 |
| Silage grass | 50 m | 3.3 cm | 3.5.2022 | 74 |
| Silage grass | 50 m | 3.3 cm | 10.6.2023 | 77 |
| Silage grass | 50 m | 3.3 cm | 11.7.2023 | 74 |
| Silage grass | 50 m | 3.3 cm | 21.6.2023 | 73 |
| Silage grass | 50 m | 3.3 cm | 24.5.2023 | 79 |
| Silage grass | 50 m | 3.3 cm | 4.6.2023 | 80 |
| Silage grass | 50 m | 3.3 cm | 5.8.2023 | 75 |
| Barley | 50 m | 3.3 cm | 20.6.2023 | 90 |
| Barley | 50 m | 3.3 cm | 20.8.2023 | 98 |
| Oats | 50 m | 3.3 cm | 7.9.2023 | 65 |
| Rapeseed | 50 m | 3.3 cm | 9.6.2023 | 141 |
| Rapeseed | 50 m | 3.3 cm | 9.8.2023 | 136 |
| Barley | 50 m | 3.3 cm | 14.6.2024 | 17 |
| Barley | 50 m | 3.3 cm | 2.7.2024 | 66 |

**TABLE A2.** MEAN AND STANDARD DEVIATION (STD) PER SPECTRAL BAND ACROSS DATASETS.

| Spectral Channel | msuav500K - Mean | msuav500K - Std | Self-collected dataset - Mean | Self-collected dataset - Std | Combined dataset: msuav500k+N - Mean | Combined dataset: msuav500k+N - Std |
|---|---|---|---|---|---|---|
| Green | 0.1535 | 0.1045 | 0.1395 | 0.0903 | 0.1545 | 0.1035 |
| Red | 0.1419 | 0.1315 | 0.1395 | 0.1045 | 0.1416 | 0.1297 |
| Red-Edge | 0.3098 | 0.1470 | 0.2403 | 0.1069 | 0.3050 | 0.1456 |
| NIR | 0.3639 | 0.1746 | 0.3089 | 0.1385 | 0.3602 | 0.1729 |

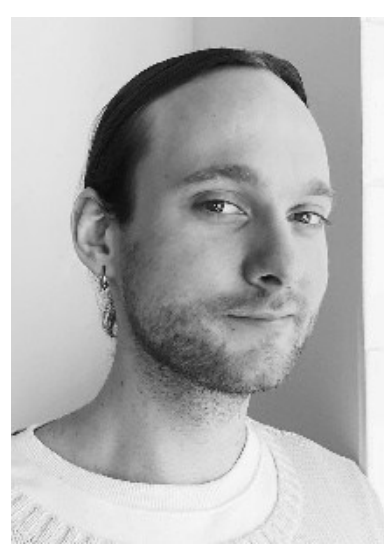

**LEON-FRIEDRICH THOMAS** received the B.Sc. degree in Environmental Sciences from Leuphana University Lüneburg in 2016 and the M.Sc. degree in Environmental Planning from the Technical University of Berlin in 2021. Since 2021, he has been pursuing a Ph.D. in Sustainable Use of Renewable Natural Resources at the University of Helsinki, where he is a member of the Agrotechnology research group within the Department of Agricultural Sciences. From 2023 to 2025, he was a visiting researcher at the Earth Observation Lab at Humboldt University of Berlin. His research focuses on remote sensing and advanced deep learning methods for Earth observation. Recently, Leon-Friedrich Thomas developed and released an open-source application for the QGIS plugin EnMAP-Box, named *Spectral Imaging Deep Learning Mapper (SpecDeepMap)*. This extension enables the efficient application of deep learning models to hyperspectral and multispectral remote sensing imagery.

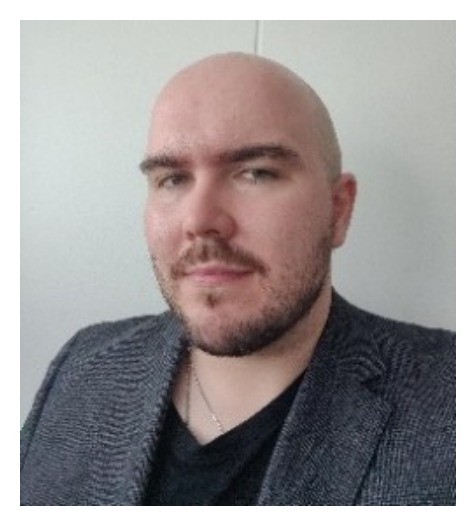

**MIKAEL ÄNÄKKÄLÄ** received the B.S. degree in Agricultural Sciences, Agrotechnology, in 2017 and the M.S. degree in Agricultural Sciences, Agrotechnology, in 2020 from the University of Helsinki, Faculty of Agriculture and Forestry, Department of Agricultural Sciences. Since 2018, he has been a Research Assistant with the same department. He is currently pursuing the Ph.D. degree in Agricultural Sciences at University of Helsinki. His doctoral research focuses on multispectral imaging of agricultural fields, with particular emphasis on the sensitivity analysis and performance evaluation of multispectral imaging systems for agricultural applications.

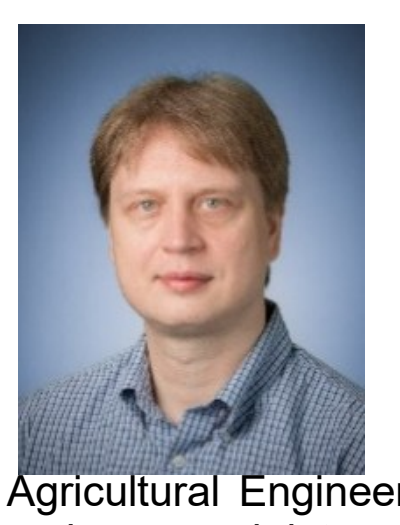

**ANTTI LAJUNEN (Senior Member, IEEE)** received the M.Sc. degree in Mechanical Engineering from Helsinki University of Technology, Finland, in 2005 and Master of Advanced Studies degree in Industrial Engineering from Ecole Centrale Paris (ECP), France, in 2007. He received his D.Sc. degree in 2014 from Aalto University, Finland. He is currently working as an Associate Professor in Agricultural Engineering at the University of Helsinki, Finland. His main research interests are electrification of agricultural vehicles and machinery, agricultural robots, automation in agriculture, and high-fidelity modelling of off-road vehicles.